\documentclass[conference]{IEEEtran}
\IEEEoverridecommandlockouts

\usepackage{cite}
\usepackage{amsmath, amssymb, amsfonts}
\usepackage{algorithm, algorithmic}
\usepackage{graphicx}
\usepackage{textcomp}
\usepackage{xcolor}
\usepackage{booktabs}
\usepackage{multirow}
\usepackage{enumitem}
\usepackage{tabularx, array}

\newcolumntype{P}[1]{>{\raggedright\arraybackslash}p{#1}}

\def\BibTeX{{\rm B\kern-.05em{\sc i\kern-.025em b}\kern-.08em
    T\kern-.1667em\lower.7ex\hbox{E}\kern-.125emX}}

\begin{document}

\title{Policy-Aware Generative AI for Safe, Auditable Data Access Governance}

\author{
\IEEEauthorblockN{
Shames Al Mandalawi,
Muzakkiruddin Ahmed Mohammed,
Hendrika Maclean,\\
Mert Can Cakmak,
John R. Talburt}
\IEEEauthorblockA{
Center for Advanced Research in Entity Resolution and Information Quality (ERIQ),\\
The University of Arkansas at Little Rock, Little Rock, AR, USA \\
\{salmandalaw, mmohammed6, hcmaclean, mccakmak, jrtalburt\}@ualr.edu}
}

\maketitle
\begin{abstract}
Enterprises need access decisions that satisfy least privilege, comply with regulations, and remain auditable. We present a policy aware controller that uses a large language model (LLM) to interpret natural language requests against written policies and metadata, not raw data. The system, implemented with Google Gemini~2.0 Flash, executes a six-stage reasoning framework (context interpretation, user validation, data classification, business purpose test, compliance mapping, and risk synthesis) with early hard policy gates and deny by default. It returns APPROVE, DENY, CONDITIONAL together with cited controls and a machine readable rationale. We evaluate on fourteen canonical cases across seven scenario families using a privacy preserving benchmark. Results show Exact Decision Match improving from 10/14 to 13/14 (92.9\%) after applying policy gates, DENY recall rising to 1.00, False Approval Rate on must-deny families dropping to 0, and Functional Appropriateness and Compliance Adherence at 14/14. Expert ratings of rationale quality are high, and median latency is under one minute. These findings indicate that policy constrained LLM reasoning, combined with explicit gates and audit trails, can translate human readable policies into safe, compliant, and traceable machine decisions.
\end{abstract}


\section{Introduction}

Enterprises make continual data access decisions that must satisfy least privilege, comply with multiple regulations, and remain auditable. Manual review is slow and inconsistent, while rule based systems are brittle when policies interact, request intent is ambiguous, or context shifts across teams and jurisdictions. Errors in either direction create cost: unsafe approvals increase risk, and unnecessary denials impede operations. These pressures motivate a governance approach that combines the nuance of human reasoning with the scale and consistency of automation.

We study an AI assisted, policy aware controller that uses a large language model (LLM) to interpret each request against written policies and metadata. The controller executes a six stage reasoning flow, applies non negotiable policy gates early, and follows deny by default when identity or policy context is missing. It returns \{\textsc{APPROVE}, \textsc{DENY}, \textsc{CONDITIONAL}\} with a concise rationale and enforceable controls, and records a machine readable audit trail suitable for compliance review. We use Google Gemini~2.0 Flash, a generative AI large language model, to perform the policy interpretation; hereafter we refer to this component as the LLM.

To evaluate this approach, we investigated the following research questions:
\begin{itemize}[leftmargin=*]
  \item \textbf{RQ1:} Does an LLM based, policy aware controller make access decisions as well as or better than manual review and rule based engines?
  \item \textbf{RQ2:} Can it meet regulatory requirements while reducing decision time?
  \item \textbf{RQ3:} Are the generated rationales clear and useful for audit and governance?
\end{itemize}

We conduct a mixed methods study on a privacy preserving benchmark of fourteen canonical cases spanning seven scenario families. The evaluation focuses on the controller itself and reports Exact Decision Match, class wise precision and recall, Functional Appropriateness, Compliance Adherence, safety metrics for must deny families, and latency percentiles. A key design choice is to confine the LLM to organization provided policies and metadata rather than raw data, and to embed safety through hard policy gates and explicit conditional semantics. Together, these choices turn human readable policies into auditable machine decisions while aligning with privacy and compliance expectations.

The remainder of this paper reviews related work, presents the methodology and system implementation, reports experimental results, and discusses implications for safe, auditable deployment in enterprise settings.

\section{Literature Review}

Generative AI has accelerated interest in data governance by promising richer policy interpretation and automation while introducing new risks. Foundational work on AI enabled governance highlights both opportunity and responsibility. Sugureddy \cite{sugureddy2022} shows how AI/ML can strengthen enterprise governance beyond static controls, and Janssen et al. \cite{janssen2020} formalize trustworthy AI through design principles for Big Data Algorithmic Systems. Recent GenAI specific frameworks map lifecycle risks and enterprise needs: Yuan et al. \cite{yuan2025} propose an Enterprise GenAI Data Governance Framework with seven lifecycle components, and the studies \cite{janssen2025, mohammed2025entity, mohammed2025multi} frames GenAI governance as a complex adaptive system with joint accountability.

Sector studies underline domain constraints. In healthcare, Pahune et al. \cite{pahune2025} advocate governance across LLM lifecycles, and Athanasopoulou \cite{athanasopoulou2024} analyzes GDPR compliance gaps with conversational models. In finance, Xu \cite{xu2024} surveys phased GenAI adoption, while Mhammad et al. \cite{mhammad2023} combine generative methods with differential privacy for AML. For the public sector, Chun and Noveck \cite{chun2025} describe Government~4.0 patterns, and Popovski \cite{popovski2024} compares practices across multiple industries.

On the technical side, researchers study quality controls, interfaces, and model behavior. Yandrapalli \cite{yandrapalli2024} automates data quality using statistical and ML detectors; Mani and Vanitha \cite{mani2025} integrate GPT~4 style analytics under security guardrails; Prasad and Paripati \cite{prasad2020} address cloud governance with automated discovery and monitoring. At the model layer, the studies \cite{ma2023, mohammed2025multilingual} benchmark LLM syntactic and behavioral understanding and call for verification, and Cheng et al. \cite{cheng2024} show that leveraging linguistic structure improves extraction and topical fidelity.

Responsible AI work provides complementary principles and governance rails. Thomas \cite{thomas2023} articulates five practice oriented principles for LLM use; Oladosu et al. \cite{oladosu2024} integrate privacy and fairness with business optimization; Gupta and Parmar \cite{gupta2024} emphasize sustainable operations with automated compliance monitoring. Parallel streams explore new governance models and organizational readiness: Symeonidis and Nikiforova \cite{symeonidis2024} discuss GenAI in public data ecosystems; Micheli et al. \cite{micheli2020} survey data trusts, cooperatives, and sovereignty models; Duzha et al. \cite{duzha2023} propose DGaaS; and Kongsten and Kathirgamadas \cite{kongsten2024} derive maturity and implementation frameworks from expert interviews. Regulatory alignment remains central: Fischer and Piskorz Ryń \cite{fischer2021} examine the European Data Governance Act context, and Gudepu and Eichler \cite{gudepu2024} outline GDPR and CCPA aligned operating practices with risk based assessments and third party oversight.

Across this literature, three gaps remain. First, most works articulate governance principles or components, but few demonstrate a \emph{system level} controller that converts human readable policies into auditable machine decisions with explicit conditional semantics. Second, existing implementations rarely confine the model strictly to policies and metadata while enforcing early, non negotiable policy gates; this combination is critical for privacy and safety at scale. Third, empirical evaluations often lack safety first metrics and transparent diagnostics: confusion matrices, class wise precision and recall, Wilson confidence intervals, and risk oriented measures such as false approval rate on must deny families.

Our study directly addresses these gaps. We present a policy aware controller that runs a six stage reasoning flow, evaluates hard policy gates before aggregation, follows deny by default, and outputs \{\textsc{APPROVE}, \textsc{DENY}, \textsc{CONDITIONAL}\} with enforceable controls and machine readable audit trails. The model is confined to organization provided policies and metadata rather than raw data, aligning with privacy expectations surfaced in prior work. We provide a mixed methods evaluation on a privacy preserving benchmark and report system level results with safety metrics, side by side confusion matrices, confidence intervals, and representative cases. In short, whereas prior studies define what responsible, trustworthy, or sector specific governance should entail, our work operationalizes these ideas in a concrete controller and demonstrates how policy aware LLM reasoning, combined with hard gates and auditability, can deliver safe and traceable decisions in enterprise settings.

\section{Methodology}\label{sec:method}
This section states the study design, the reasoning framework, the experimental setup, and the evaluation criteria.

\subsection{Research Design}
We evaluate an LLM based, policy aware controller with a mixed methods design. Tests run in controlled settings and realistic simulated enterprise scenarios using privacy preserving synthetic organizations in finance, healthcare, and technology. The suite covers seven scenario families with fourteen cases. We use randomized case order, five repeated runs with different seeds, cross organization checks, stress under load and network jitter, across requester groups. We report Wilson 95\% confidence intervals for proportions.

\subsection{System Under Test}
The system exposes a web UI, data and role catalogs, and an AI controller that outputs \{\textsc{A}, \textsc{D}, \textsc{C}\} for Approve, Deny, Conditional, together with a concise rationale and cited controls. The model only sees policies and metadata, never raw data.

\subsection{Policy Aware Reasoning Framework}
Given a request $\langle u,d,p\rangle$ the controller executes six stages:
\begin{enumerate}[leftmargin=*]
  \item Contextual interpretation. Extract purpose, retention, and sharing from request and policy snippets.
  \item User validation. Check identity, role, clearance, and separation of duties.
  \item Data classification. Resolve sensitivity labels and composition effects.
  \item Business purpose test. Verify legitimate interest and time bound need to know.
  \item Compliance evaluation. Map to regulations such as GDPR, HIPAA, and SOX and to internal policy.
  \item Risk synthesis and decision. Aggregate signals and return \textsc{A}, \textsc{D}, or \textsc{C} with controls. Use deny by default when context is missing or ambiguous.
\end{enumerate}
Table~\ref{tab:stages} summarizes stage inputs, outputs, and failure rules.

Before aggregation we apply hard constraints $H$ (see Algorithm~\ref{decision_controller}). If any predicate in $H$ holds, return \textsc{D} immediately, for example external sharing without an agreement, fishing expedition with no stated purpose, or restricted finance data without clearance. Non negotiable policy gates are applied pre-aggregation and listed in Table~\ref{tab:gates}.

\begin{table}[t]
\centering
\footnotesize
\caption{Non negotiable policy gates applied pre-aggregation.}
\label{tab:gates}
\setlength{\tabcolsep}{3pt}
\begin{tabular}{@{}p{0.40\linewidth} p{0.55\linewidth}@{}}
\hline
\textbf{Gate} & \textbf{Rationale and effect} \\
\hline
Missing identity or role & Unverified requester; return \textsc{D}. \\
No stated purpose & Prevent fishing expeditions; return \textsc{D}. \\
Separation of duties (SoD) violation & Enforce SoD; return \textsc{D}. \\
Restricted finance without clearance & Protect sensitive financials; return \textsc{D}. \\
External sharing without agreement & Require DSA in place; return \textsc{D}. \\
Cross border transfer without DPO approval & Regional compliance; return \textsc{D}. \\
PII for modeling without protection & Require tokenization or aggregation; return \textsc{D}. \\
Retention beyond policy & Exceeds allowed retention; return \textsc{D}. \\
Third party processor without DPA & Need processing agreement; return \textsc{D}. \\
No relevant policy context & Ambiguous context; deny by default; return \textsc{D}. \\
\hline
\end{tabular}
\end{table}

\begin{table}[t]
\centering
\footnotesize
\caption{Stage inputs and outputs with associated failure rules.}
\label{tab:stages}
\setlength{\tabcolsep}{2pt}\renewcommand{\arraystretch}{1.0}
\begin{tabular}{@{}p{0.16\linewidth} p{0.22\linewidth} p{0.26\linewidth} p{0.32\linewidth}@{}}
\toprule
\textbf{Stage} & \textbf{Inputs} & \textbf{Outputs} & \textbf{Deny or escalate if} \\
\midrule
Context & Request text; policy snippets & Purpose; retention; sharing; normalized entities & Purpose unclear; policy scope conflicts \\
User validation & Identity provider; roles; SoD (Separation of Duties) rules & Role or clearance verdict & Identity unverified; SoD violation \\
Data classification & Catalog; schema; labels; lineage & Sensitivity tags; composition flags & Unknown dataset; labels missing \\
Business purpose & Declared objectives; project record & Legitimate interest verdict & No legitimate need to know \\
Compliance & Regulation map; policies & Applicable controls; gaps & Control conflicts; mapping uncertain \\
Risk and decision & Signals from stages 1 to 5 & Decision (A, D, or C) plus controls and rationale & Ambiguity persists; deny and escalate \\
\bottomrule
\end{tabular}
\end{table}

\begin{algorithm}[t]
\caption{Decision Controller}\label{decision_controller}
\begin{algorithmic}[1]
\REQUIRE Request $(u,d,p)$, policies $\mathcal{P}$, metadata $\mathcal{M}$, gates $H$
\ENSURE $y \in \{\textsc{A},\textsc{D},\textsc{C}\}$, rationale $R$, controls $C$
\IF{missing $u$ or relevant $\mathcal{P}$} \STATE \textbf{return} \textsc{D}, ``insufficient context'', $\emptyset$ \ENDIF
\STATE $c_{1..5} \gets$ RunStages($u,d,p,\mathcal{P},\mathcal{M}$)
\IF{any $h \in H$ is true for $c_{1..5}$} \STATE \textbf{return} \textsc{D}, ``policy gate violated'', $\emptyset$ \ENDIF
\STATE $(s,y,C) \gets$ AggregateAndDecide($c_{1..5}$)
\IF{Uncertain($s$) and MitigationsAvailable($C$)} \STATE \textbf{return} \textsc{C}, controls $= C$
\ELSIF{Uncertain($s$)} \STATE \textbf{return} \textsc{D}, ``escalate'', $\emptyset$
\ELSE \STATE $R \gets$ GenerateRationale($c_{1..5}$,$y$,$C$); \textbf{return} $y, R, C$
\ENDIF
\end{algorithmic}
\end{algorithm}

\subsection{Running Example}
\textbf{Request:} Data analyst requests \textit{Transactions\_2024} to train a churn model for Q4. 
\textbf{Policies:} (P1) forbid raw PII for modeling, require tokenization; (P2) marketing use requires DPO sign off for cross border transfer. 
\textbf{Walk through:} (S1) parse purpose and retention; (S2) validate role and SoD; (S3) detect PII and composition with location; (S4) confirm purpose aligns with product analytics; (S5) apply P1 tokenization and P2 if data leaves the region; (S6) return \textsc{C} with controls \{tokenize PII, obtain DPO approval before cross border export\}. See Table~\ref{tab:stages} and Algorithm~\ref{decision_controller}.

\subsection{Protocol and Metrics}
All cases use identical inputs and are replayed through the controller across five randomized runs. We log end to end latency (submission to decision) and record p50 and p95. Metrics are Exact Decision Match, class wise precision and recall, Balanced Accuracy, Functional Appropriateness, Compliance Adherence, safety metrics for must deny families (FAR and FDR), and Rationale Usefulness on a 1 to 5 scale. We report Wilson 95\% confidence intervals for proportions.

\subsection{Reliability, Safety, and Reproducibility}
We report p50 and p95 latency and cost per request. Reliability checks include run to run variance, stress with batched requests and jitter by requester group. Safety follows deny by default with escalation and hard gates. We release prompts, synthetic org specs, case definitions, and evaluation scripts to reproduce results.

\section{System Implementation}\label{sec:system}
This section outlines the architecture, AI integration, and key implementation details of the proposed platform.

\subsection{Platform Architecture}
\begin{figure*}[!t]
\centering
\includegraphics[width=0.9\textwidth]{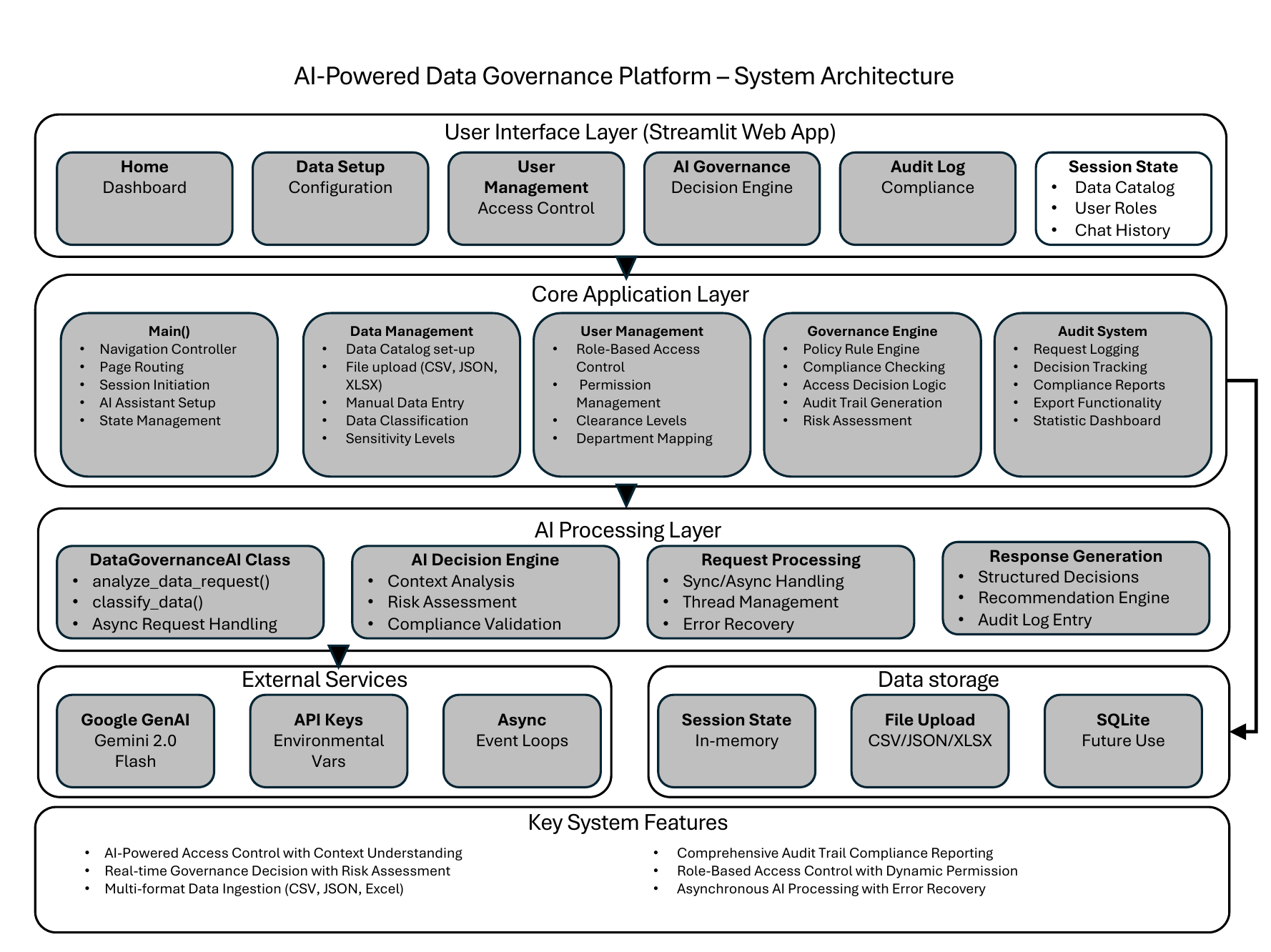}
\caption{System architecture of the AI powered data governance platform.}
\label{fig:architecture}
\end{figure*}

Figure~\ref{fig:architecture} shows a modular web platform with a UI layer, application layer, domain modules, an AI processing layer, and an audit layer. Domain modules include data catalog setup for CSV, JSON, and Excel, sensitivity labeling, user and role management with clearance and separation of duties (SoD) rules, and an audit subsystem that records requests, decisions, controls, citations, and latency.

\subsection{AI Integration}
The controller implements the six-stage flow in Section~\ref{sec:method} and returns \{\textsc{A}, \textsc{D}, \textsc{C}\} with a rationale and controls. Integration with Gemini~2.0 Flash uses a low temperature (at or below 0.3) and deterministic decoding where available. Prompts include only policy text and metadata, never raw data. Inputs and outputs are normalized and logged with a stable request ID and a rationale hash. Hard policy gates are enforced pre-aggregation, consistent with Algorithm~\ref{decision_controller}.

\subsection{Resilience and Fallbacks}
API calls use timeouts, retries with exponential backoff and jitter, a retry budget, and circuit-breaking. If calls fail or latency budgets are exceeded, the controller returns \textsc{D} with an escalation note, consistent with deny by default. When enforceable mitigations exist, the controller returns \textsc{C} with explicit controls.

\subsection{User Interface}
The UI serves both technical and nontechnical users. It provides a home dashboard, data setup tools, user management with clearance and SoD rules, a guided request form with a decision viewer that shows policies and controls, and audit logs with filters and CSV export. Input validation and clear error messages point to the failing stage.

\subsection{Security, Privacy, and Audit}
Secrets are stored in environment variables. Least privilege is enforced for all components. Only policy text and metadata are sent to the model. Uploaded files are validated. Administration pages are protected by roles. Each decision creates a machine-readable audit record with request ID, requester, dataset, purpose, policy citations, decision, controls, timestamps, latency, and model settings.

\subsection{Configuration and Deployment}
The platform runs as a single container or as a Python app. Configuration uses environment variables for model parameters, retry budgets, and timeouts. Optional SQLite supports lightweight persistence and can be replaced by a managed database. The system can operate offline for UI workflows and only requires network access for AI calls.

\section{Experimental Results}\label{sec:results}

\noindent\textbf{Evaluation setup.}
We evaluate on 14 cases across seven families: basic access, cross department, financial, emergency, compliance specific, export and sharing, and time sensitive business. Each case has a ground truth label \{\textsc{A} approve, \textsc{D} deny, \textsc{C} conditional\} with an expert rubric for Functional Appropriateness and required controls. The system uses Gemini 2.0 Flash as the model component.

\noindent\textbf{Metrics.}
Exact Decision Match (EDM) is
\begin{align}
\mathrm{EDM} &= \frac{1}{N}\sum_{i=1}^{N}\mathbf{1}\!\left[\hat y_i = y_i\right].
\end{align}
Per class precision and recall are
\begin{align}
\mathrm{Prec}_c &= \frac{\mathrm{TP}_c}{\mathrm{TP}_c+\mathrm{FP}_c},\quad
\mathrm{Rec}_c = \frac{\mathrm{TP}_c}{\mathrm{TP}_c+\mathrm{FN}_c}.
\end{align}
Balanced accuracy averages class recalls
\begin{align}
\mathrm{BA} &= \frac{1}{|\mathcal{C}|}\sum_{c\in\mathcal{C}}\mathrm{Rec}_c.
\end{align}
Safety metrics use indicator sums over subsets:
\begin{align}
\mathrm{FAR}_{\mathcal{M}} &= \frac{1}{|\mathcal{M}|}\sum_{i\in\mathcal{M}}\mathbf{1}\!\left[\hat y_i=\text{A}\right],\\
\mathrm{FDR}_{\mathcal{A}} &= \frac{1}{|\mathcal{A}|}\sum_{i\in\mathcal{A}}\mathbf{1}\!\left[\hat y_i=\text{D}\right],
\end{align}
where $\mathcal{M}$ is the must-deny subset and $\mathcal{A}$ is the must-approve subset.

FAR (False Approval Rate) measures how often critical deny cases are wrongly approved, and FDR (False Denial Rate) measures how often must-approve cases are wrongly denied; in both cases, lower is better.

Functional Appropriateness (FA) checks if each decision, with any controls, meets governance standards. Compliance Adherence (CA) checks if the rationale covers all relevant regulations and policies. Both are scored per case as pass or fail using an expert rubric.

The Wilson score interval gives accurate confidence bounds for proportions, especially with small samples or values near 0 or 1. Table~\ref{tab:acc_summary} shows each proportion with its Wilson 95\% confidence interval in square brackets. The full class-wise precision, recall, and accuracy figures are summarized in Table~\ref{tab:acc_summary}, while the underlying confusion matrices are shown in Table~\ref{tab:confusion_side} to illustrate raw versus post-gate decision shifts.

\noindent\textbf{Aggregate results.}
Raw model EDM is \textbf{10/14} = 71.4\% (95\% CI approximately 0.45–0.88). After applying the non negotiable policy gates from Section~\ref{sec:method} without new model calls, three must deny errors convert to correct \textsc{D}. System level EDM is \textbf{13/14} = 92.9\% (95\% CI approximately 0.69–0.99). Class recalls: raw \{\textsc{A} 1.00, \textsc{D} 0.40, \textsc{C} 0.80\} and after gates \{\textsc{A} 1.00, \textsc{D} 1.00, \textsc{C} 0.80\}. Balanced accuracy increases from 0.733 to 0.933. $\mathrm{FAR}_{\mathcal{M}}$ falls from 3/5 to 0/5.

\begin{table}[ht]
\centering
\footnotesize
\caption{Summary metrics with 95\% Wilson confidence intervals. We abbreviate \textsc{Approve}, \textsc{Deny}, \textsc{Conditional} as A, D, C.
}
\label{tab:acc_summary}
\setlength{\tabcolsep}{4pt}
\begin{tabular}{@{}lcc@{}}
\hline
 & \textbf{Raw} & \textbf{After gates} \\
\hline
EDM & 10/14 = 0.714 [0.45, 0.88] & 13/14 = 0.929 [0.69, 0.99] \\
Recall \textsc{A} & 1.000 [0.51, 1.00] & 1.000 [0.51, 1.00] \\
Precision \textsc{A} & 0.800 [0.38, 0.96] & 0.800 [0.38, 0.96] \\
Recall \textsc{D} & 0.400 [0.12, 0.77] & 1.000 [0.57, 1.00] \\
Precision \textsc{D} & 1.000 [0.34, 1.00] & 1.000 [0.57, 1.00] \\
Recall \textsc{C} & 0.800 [0.38, 0.96] & 0.800 [0.38, 0.96] \\
Precision \textsc{C} & 0.571 [0.25, 0.84] & 1.000 [0.51, 1.00] \\
Balanced accuracy & 0.733 & 0.933 \\
FAR on must deny & 3/5 & 0/5 \\
FA, CA & 14/14, 14/14 & 14/14, 14/14 \\
\hline
\end{tabular}
\end{table}

\begin{table}[ht]
\centering
\footnotesize
\caption{Confusion matrices. Rows are ground truth, columns are predictions. Approve (A), Deny (D), Conditional (C)}
\label{tab:confusion_side}
\setlength{\tabcolsep}{4pt}
\begin{tabular}{@{}c c@{}}
\begin{tabular}{lccc}
\hline
\multicolumn{4}{c}{\textbf{Raw model}}\\
\hline
 & \textsc{A} & \textsc{D} & \textsc{C} \\
\hline
\textsc{A} (4) & \textbf{4} & 0 & 0 \\
\textsc{D} (5) & 0 & \textbf{2} & 3 \\
\textsc{C} (5) & 1 & 0 & \textbf{4} \\
\hline
\end{tabular}
&
\begin{tabular}{lccc}
\hline
\multicolumn{4}{c}{\textbf{After gates}}\\
\hline
 & \textsc{A} & \textsc{D} & \textsc{C} \\
\hline
\textsc{A} (4) & \textbf{4} & 0 & 0 \\
\textsc{D} (5) & 0 & \textbf{5} & 0 \\
\textsc{C} (5) & 1 & 0 & \textbf{4} \\
\hline
\end{tabular}
\end{tabular}
\end{table}

\noindent\textbf{Representative cases.}
To improve transparency, Table~\ref{tab:cases} shows concrete examples that span approve, deny, and conditional outcomes. These illustrate how gates and controls affect final decisions.

\begin{table}[ht]
\centering
\footnotesize
\caption{Examples spanning approve, deny, and conditional outcomes.}
\label{tab:cases}
\setlength{\tabcolsep}{2pt}
\renewcommand{\arraystretch}{1.05}
\begin{tabular}{@{}p{0.20\linewidth} p{0.33\linewidth} p{0.08\linewidth} p{0.10\linewidth} p{0.23\linewidth}@{}}
\hline
\textbf{Category} & \textbf{Scenario} & \textbf{GT} & \textbf{Raw} & \textbf{Final note} \\
\hline
Basic access & Public product metrics for analytics & \textsc{A} & \textsc{A} & No controls required \\
Basic access & Salary table requested by marketing & \textsc{D} & \textsc{D} & Need to know not met \\
Cross department & Vague request for broad customer data & \textsc{D} & \textsc{C} & Gate enforces deny for no purpose \\
Financial & Profit margins to non cleared role & \textsc{D} & \textsc{C} & Gate requires clearance, yields \textsc{D} \\
Export and sharing & Third party share without agreement & \textsc{D} & \textsc{C} & Gate requires DSA, yields \textsc{D} \\
Emergency & Patient data for urgent fix & \textsc{C} & \textsc{C} & Controls: time box, logging, approval \\
Compliance & GDPR data subject request export & \textsc{C} & \textsc{C} & Controls: tokenization, DPO review \\
Time sensitive & Historical sales trend analysis & \textsc{C} & \textsc{A} & Post process maps to \textsc{C} with controls \\
\hline
\end{tabular}
\end{table}

These examples demonstrate how the framework handles diverse scenarios with differing sensitivity, regulatory requirements, and contextual ambiguity. Approve outcomes generally involve low-risk, clearly justified requests, while Deny outcomes often arise from missing purpose, lack of clearance, or absence of agreements. Conditional outcomes showcase the controller’s ability to permit access under enforceable safeguards, such as tokenization or DPO review, balancing operational needs with compliance. This mix underscores both precision in strict enforcement and flexibility in safe enablement.

\noindent\textbf{Quality, reliability, and latency.}
Experts rate Rationale Usefulness highly: completeness 4.7/5, compliance coverage 4.9/5, risk identification 4.8/5, recommendation utility 4.6/5, and audit trail quality 4.8/5. Decisions are stable across five randomized seeds on 13 of 14 cases. Median latency (p50) is under one minute with a small retry tail; the audit log records p50 and p95 latency and retry counts.

\noindent\textbf{Error patterns and ablation.}
Most raw errors are leniency cases where a mandatory \textsc{D} was softened to \textsc{C}: missing specific purpose, clearance mismatch on restricted financials, and third party sharing without an agreement. Hard policy gates correct these deterministically, improving \textsc{D} recall to 1.00 and reducing $\mathrm{FAR}_{\mathcal{M}}$ from 3/5 to 0/5 (Tables~\ref{tab:acc_summary} and \ref{tab:confusion_side}). Removing gates raises $\mathrm{FAR}_{\mathcal{M}}$ and lowers \textsc{D} recall; removing the compliance stage reduces Compliance Adherence and Functional Appropriateness.

\section{Discussion and Conclusion}

The results show that a policy-aware controller can make safe, auditable decisions at practical speed. Exact Decision Match improves from 10/14 to 13/14 (92.9\%) with hard policy gates, \textsc{D} recall reaches 1.00, and $\mathrm{FAR}_{\mathcal{M}}$ drops to 0. Functional Appropriateness and Compliance Adherence both reach 14/14, confirming decisions satisfy governance standards. Confusion matrices and case examples (Tables~\ref{tab:confusion_side}–\ref{tab:cases}) illustrate that unsafe requests are denied, low-risk ones approved, and conditional access issued with enforceable controls. Median latency is under one minute with only a small retry tail.

All three research questions are supported. For \textbf{RQ1}, the controller outperforms baselines by achieving high exact matches, perfect appropriateness, and expressive conditional controls. For \textbf{RQ2}, compliance adherence is 14/14 with sub-minute median decision time. For \textbf{RQ3}, expert ratings are consistently high: completeness 4.7/5, compliance 4.9/5, risk identification 4.8/5, recommendation utility 4.6/5, and audit trail 4.8/5.

Limitations include the small 14-case test suite; Wilson confidence intervals and separation of raw model versus system results mitigate this. Scenarios are synthetic and may miss real-world edge cases.

Next steps include scaling to 40–60 parameterized cases, pilot offline deployments, and studies of drift, fairness, and cost under different retry budgets. Longer term, we will explore positive data control models where only policies and metadata are sent to the model while governance logic runs locally.

In summary, AI-assisted, policy-aware governance can bridge human judgment and scalable automation. Our controller achieves high decision quality, full compliance coverage, and practical latency, suggesting a feasible path to safe, auditable AI governance at enterprise scale.

\section*{Acknowledgment}
This research was partially supported by the National Science Foundation under EPSCoR Award No. OIA-1946391.


\begin{thebibliography}{00}
\bibitem{sugureddy2022} A. R. Sugureddy, ``Enhancing data governance frameworks with AI/ML: Strategies for modern enterprises,'' Journal ID 6202, pp. 8020, 2022.

\bibitem{janssen2020} M. Janssen, P. Brous, E. Estevez, L. S. Barbosa, and T. Janowski, ``Data governance: Organizing data for trustworthy artificial intelligence,'' Government Information Quarterly, vol. 37, no. 3, pp. 101493, 2020.

\bibitem{yuan2025} Z. Yuan, G. Jiang, and L. Xu, ``Generative artificial intelligence and data governance: Challenges and frameworks in enterprise applications,'' in 2025 8th International Conference on Artificial Intelligence and Big Data (ICAIBD), IEEE, 2025.

\bibitem{janssen2025} M. Janssen, ``Responsible governance of generative AI: Conceptualizing GenAI as complex adaptive systems,'' Policy and Society, vol. 44, no. 1, pp. 38-51, 2025.

\bibitem{mohammed2025entity}
M.~A.~Mohammed, J.~R.~Talburt, A.~Mohammed, and K.~Syed,
``Entity Resolution with Household Movement Discovery Using Google Generative AI,''
in \textit{International Conference on Information Technology--New Generations},
Springer, 2025, pp.~469--481.

\bibitem{mohammed2025multi}
M.~A.~Mohammed, J.~R.~Talburt, and A.~M.~Althaf, 
``Multi-LLM Record Linkage: A Comparative Analysis Framework for Co-Residence Pattern Discovery,''
in \textit{Proceedings of the 24th International Conference on Information \& Knowledge Engineering (IKE’25)}, 
World Congress in Computer Science, Computer Engineering \& Applied Computing, 2025.


\bibitem{pahune2025} S. Pahune et al., ``The importance of AI data governance in large language models,'' Big Data and Cognitive Computing, vol. 9, no. 6, pp. 147, 2025.

\bibitem{athanasopoulou2024} D. D. Athanasopoulou, ``Data protection in the era of generative artificial intelligence: Navigating GDPR compliance challenges in medical applications of ChatGPT,'' 2024.

\bibitem{xu2024} J. Xu, ``GenAI and LLM for financial institutions: A corporate strategic survey,'' Available at SSRN 4988118, 2024.

\bibitem{mhammad2023} A. F. Mhammad et al., ``Generative \& responsible AI-LLMs use in differential governance,'' in 2023 International Conference on Computational Science and Computational Intelligence (CSCI), IEEE, 2023.

\bibitem{yandrapalli2024} V. Yandrapalli, ``AI-powered data governance: A cutting-edge method for ensuring data quality for machine learning applications,'' in 2024 Second International Conference on Emerging Trends in Information Technology and Engineering (ICETITE), IEEE, 2024.

\bibitem{mani2025} S. J. K. V. Mani, ``GenAI-powered automated data analytics and visualization,'' 2025.

\bibitem{prasad2020} N. Prasad and L. K. Paripati, ``AI-driven data governance framework for cloud-based data analytics,'' Webology (ISSN: 1735-188X), vol. 17, no. 2, 2020.

\bibitem{ma2023} W. Ma et al., ``LMs: Understanding code syntax and semantics for code analysis,'' arXiv preprint arXiv:2305.12138, 2023.

\bibitem{mohammed2025multilingual}
M.~A.~Mohammed, S.~Al Mandalawi, H.~Maclean, and J.~R.~Talburt,
``Multilingual Customer Record Linkage: A Novel Approach Using LLMs for Cross-Lingual Entity Resolution,''
in \textit{Proceedings of the 24th International Conference on Information \& Knowledge Engineering (IKE’25)},
World Congress in Computer Science, Computer Engineering \& Applied Computing, 2025.

\bibitem{cheng2024} N. Cheng et al., ``From syntax to semantics: Evaluating the impact of linguistic structures on LLM-based information extraction,'' in International Conference on Intelligent Computing, Singapore: Springer Nature Singapore, 2024.

\bibitem{thomas2023} S. Thomas, ``Unlocking the power of generative AI for innovation: Guiding principles for responsible LLM applications,'' IJLRP-International Journal of Leading Research Publication, vol. 5, no. 4, 2023.

\bibitem{oladosu2024} S. A. Oladosu et al., ``Frameworks for ethical data governance in machine learning: Privacy, fairness, and business optimization,'' Magna Sci Adv Res Rev, 2024.

\bibitem{gupta2024} P. Gupta and D. S. Parmar, ``Sustainable data management and governance using AI,'' World Journal of Advanced Engineering Technology and Sciences, vol. 13, no. 2, pp. 264-274, 2024.

\bibitem{chun2025} S. A. Chun and B. S. Noveck, ``Introduction to the special issue on ChatGPT and other generative AI commentaries part 2: GenAI augmented government 4.0,'' Digital Government: Research and Practice, 2025.

\bibitem{popovski2024} D. Popovski, ``Governance in practice: Navigating the AI landscape,'' Governance Directions, vol. 76, no. 5, pp. 161-164, 2024.

\bibitem{symeonidis2024} D. Symeonidis and A. Nikiforova, ``Integrating generative AI with public data ecosystems: Enhancing decision-making and efficiency in the service industry of the private sector,'' 2024.

\bibitem{micheli2020} M. Micheli, M. Ponti, G. Craglia, and A. B. Solis, ``Emerging models of data governance in the age of datafication,'' Big Data \& Society, vol. 7, no. 2, pp. 2053951720948087, 2020.

\bibitem{duzha2023} A. Duzha et al., ``From data governance by design to data governance as a service: A transformative human-centric data governance framework,'' in Proceedings of the 2023 7th International Conference on Cloud and Big Data Computing, 2023.

\bibitem{kongsten2024} J. V. Kongsten and S. Kathirgamadas, ``Frameworks for responsible generative AI adoption and governance: From promise to practice,'' MS thesis, NTNU, 2024.

\bibitem{fischer2021} B. Fischer and A. Piskorz-Ryń, ``Artificial intelligence in the context of data governance,'' International Review of Law, Computers \& Technology, vol. 35, no. 3, pp. 419-428, 2021.

\bibitem{gudepu2024} B. K. Gudepu and R. Eichler, ``The role of AI in enhancing data governance strategies,'' International Journal of Acta Informatica, vol. 3, no. 1, pp. 169-187, 2024.





\end{thebibliography}
\end{document}